# Contextual Weak Independence in Bayesian Networks

S.K.M. Wong
Department of Computer Science
University of Regina
Regina, Saskatchewan
Canada S4S0A2
wong@cs.uregina.ca

C.J. Butz
Department of Computer Science
University of Regina
Regina, Saskatchewan
Canada S4S0A2
butz@cs.uregina.ca

## Abstract

It is well-known that the notion of (strong) *conditional independence* (CI) is too restrictive to capture independencies that only hold in certain contexts. This kind of contextual independency, called *context-strong independence* (CSI), can be used to facilitate the acquisition, representation, and inference of probabilistic knowledge. In this paper, we suggest the use of *contextual weak independence* (CWI) in Bayesian networks. It should be emphasized that the notion of CWI is a more general form of contextual independence than CSI. Furthermore, if the contextual strong independence holds for all contexts, then the notion of CSI becomes strong CI. On the other hand, if the weak contextual independence holds for all contexts, then the notion of CWI becomes *weak independence* (WI) which is a more general noncontextual independency than strong CI. More importantly, *complete* axiomatizations are studied for both the class of WI and the class of CI and WI together. Finally, the interesting property of WI being a necessary and sufficient condition for ensuring *consistency* in granular probabilistic networks is shown.

## 1 INTRODUCTION

In the probabilistic approach to uncertainty management, one assumes that knowledge can be represented as a *joint probability distribution* [6]. In practice it may not be feasible to elicit and store the required probability values when the number of variables becomes large. However, we can utilize the notion of (strong) *conditional independence* (CI) to economically represent a joint probability distribution as the product of *conditional probability tables* (CPTs). The required joint probability values can then be obtained indirectly by eliciting the corresponding conditional probabilities.

Many researchers including [1, 2, 3, 4, 6] have pointed out that the notion of strong CI is too restrictive to capture independencies that hold in some but not necessarily all contexts. This contextual strong independency, referred to as *context-strong independence* (CSI) [1], *asymmetric independence* (ASI) [3], or *probabilistic causal irrelevance* (PCI) [2], can be used to facilitate the acquisition, representation, and inference of probabilistic knowledge. Geiger and Heckerman [3] proposed the use of multiple Bayesian networks (*Bayesian multinets*) to represent and reason with contextual strong independence statements. That is, each Bayesian network reflects the CI statements that hold for a given context. The particular Bayesian network used during inference is determined by the evidence. An alternative method suggested by Boutilier et al. [1], is to incorporate auxiliary nodes into a single Bayesian network. In this case, additional nodes are used to reflect the strong conditional independence statements that hold in a given context. In [2], on the other hand, Galles and Pearl develop axioms for inferring contextual strong independencies in a similar spirit to inferring strong conditional independencies using the semi-graphoid axioms [6].

In this paper, we suggest the use of *contextual weak independence* (CWI) in Bayesian networks. It should be emphasized that the notion of CWI is a more general form of contextual independence than CSI, ASI and PCI. If the contextual strong independence holds for all contexts, then the notions of CSI, ASI and PCI become strong CI. On the other hand, if the weak contextual independence holds for all contexts, then the notion of CWI becomes *weak independence* (WI). Just as contextual weak independence is more general than the notions of contextual strong independence, it is explicitly demonstrated that the notion of WI is a more general noncontextual independency than strong CI. More importantly, *complete* axiomatizations are stud-ied for both the class of WI and the class of CI and WI together. Finally, the interesting property of WI being a necessary and sufficient condition for ensuring *consistency* in granular probabilistic networks is shown. We use the term granular to mean the ability to *coarsen* and *refine* parts of a probabilistic network depending on whether they are of interest or not [5].

This paper is organized as follows. Section 2 contains background knowledge. In Section 3, we introduce the notion of CWI. The noncontextual independency WI is presented in Section 4. In Section 5, a *complete* axiomatization is shown for both the class of WI only and the class of CI and WI together. The application of WI to granular probabilistic networks is discussed in Section 6. The conclusion is presented in Section 7.

## 2 BASIC NOTIONS

In this section, we review pertinent notions including (strong) probabilistic conditional independence and the more general notion of contextual strong independence [1, 2, 3].

Consider a finite set $\mathbf{U} = \{A_1, A_2, \ldots, A_n\}$ of discrete random variables, where each variable $A \in \mathbf{U}$ takes on values from a finite domain $V_A$. We may use capital letters, such as $X, Y, Z$, for variable names and lowercase letters $x, y, z$ to denote specific values taken by those variables. Sets of variables will be denoted by boldfaced capital letters $\mathbf{X}, \mathbf{Y}, \mathbf{Z}$, and assignments of values to the variables in these sets (called configurations or tuples) will be denoted by boldfaced lowercase letters $\mathbf{x}, \mathbf{y}, \mathbf{z}$. We use $V_\mathbf{X}$ in the obvious way. We shall also use the short notation $P(x)$ for the probabilities $P(X = x)$, $x \in V_X$, and $P(\mathbf{z})$ for the set of variables $\mathbf{Z} = \{X, Y\} = XY$ meaning

$$P(\mathbf{Z} = \mathbf{z}) = P(X = x, Y = y) = P(x, y),$$

where $x \in V_X, y \in V_Y$.

Let $P$ be a joint probability distribution over the variables in $\mathbf{U}$ and $\mathbf{X}, \mathbf{Y}, \mathbf{Z}$ be subsets of $\mathbf{U}$. We say $\mathbf{X}$ and $\mathbf{Z}$ are *conditionally independent* given $\mathbf{Y}$, denoted $I(\mathbf{X} \perp \mathbf{Z} \mid \mathbf{Y})$, if given any $\mathbf{x} \in V_\mathbf{X}, \mathbf{y} \in V_\mathbf{Y}$, then for all $\mathbf{z} \in V_\mathbf{Z}$:

$$P(\mathbf{x} \mid \mathbf{y}, \mathbf{z}) = P(\mathbf{x} \mid \mathbf{y}), \text{ whenever } P(\mathbf{y}, \mathbf{z}) > 0. \quad (1)$$

For convenience we write equation (1) as $P(\mathbf{X} \mid \mathbf{Y}, \mathbf{Z}) = P(\mathbf{X} \mid \mathbf{Y})$. Alternatively, the same strong conditional independence (CI) can be defined as

$$P(\mathbf{x}, \mathbf{y}, \mathbf{z}) = \frac{P(\mathbf{x}, \mathbf{y}) \cdot P(\mathbf{y}, \mathbf{z})}{P(\mathbf{y})}, \quad (2)$$

where $P(\mathbf{x}, \mathbf{y}), P(\mathbf{y}, \mathbf{z})$ and $P(\mathbf{y})$ are *marginal* distributions of $P(\mathbf{x}, \mathbf{y}, \mathbf{z})$.

The notion of strong CI is extensively used economically representing a joint probability distribution as a Bayesian network. A *Bayesian network* [6] is a directed acyclic graph (DAG) together with a corresponding set of conditional probability tables. By definition, a Bayesian network only reflects conditional independencies $P(\mathbf{x} \mid \mathbf{y}, \mathbf{z}) = P(\mathbf{x} \mid \mathbf{y})$ which hold for *all* $\mathbf{y} \in V_\mathbf{Y}$. In some situations, however, the conditional independence may only hold for certain *specific* values in $V_\mathbf{Y}$. For example, consider the CPT depicted in Table 1. It can be seen that variables $X$ and $\{Z, W\}$ are strongly *conditionally independent* given the *context* $Y = 0$. However, $X$ and $\{Z, W\}$ are *not* strongly conditionally independent given the context $Y = 1$ since

$$P(X = 1 \mid Y = 1, Z = 0, W = 0)$$
$$= p_3 \neq P(X = 1 \mid Y = 1, Z = 1, W = 1) = p_4.$$

Table 1: Variables $X$ and $\{Z, W\}$ are strongly *conditionally independent* given the *context* $Y = 0$, but *not* when $Y = 1$.

| X | Y | Z | W | $P(X \mid Y, Z, W)$ |
|---|---|---|---|---|
| 0 | 0 | 0 | 0 | $p_1$ |
| 0 | 0 | 0 | 1 | $p_1$ |
| 0 | 0 | 1 | 0 | $p_1$ |
| 0 | 0 | 1 | 1 | $p_1$ |
| 1 | 0 | 0 | 0 | $p_2$ |
| 1 | 0 | 0 | 1 | $p_2$ |
| 1 | 0 | 1 | 0 | $p_2$ |
| 1 | 0 | 1 | 1 | $p_2$ |
| 0 | 1 | 0 | 0 | $p_3$ |
| 0 | 1 | 0 | 1 | $p_3$ |
| 0 | 1 | 1 | 0 | $p_3$ |
| 0 | 1 | 1 | 1 | $p_3$ |
| 1 | 1 | 0 | 0 | $p_3$ |
| 1 | 1 | 0 | 1 | $p_3$ |
| 1 | 1 | 1 | 0 | $p_4$ |
| 1 | 1 | 1 | 1 | $p_4$ |

Many different approaches including [1, 2, 3, 4, 6] have been proposed for representing and reasoning with independencies that are more general than CI. In [4, 6], the notion of *causal* independence is suggested to facilitate knowledge acquisition and increase the speed of inference. However, the work most relevant to this paper involves the notions of contextual strong independence, namely, *asymmetric independence* [3], *context-strong independence* [1], and *probabilistic causal irrelevance* [2].

Geiger and Heckerman [3] studied the notion of *asymmetric independence* (ASI), which states that variables are independent for some but not necessarily for all of their values. The use of *multiple* Bayesian networks (*Bayesian multinets*) are then proposed since contextual strong independence statements cannot be represented naturally in a Bayesian network. For example, consider again the CPT in Table 1. The Bayesian network constructed using the notion of CI does not reflect any independence between variables $X$ and $\{Z, W\}$ given $Y$. However, one Bayesian network can be constructed to capture the conditional independence of $X$ and $\{Z, W\}$ given $Y = 0$, and another to show their dependence when $Y = 1$ as shown in Figure 1 (left). The particular Bayesian network to be used in the inference process is determined by the evidence.

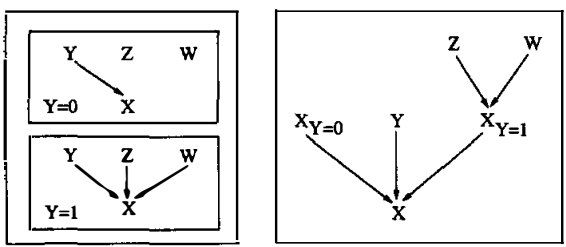

Figure 1: The use of ASI in constructing *Bayesian multinets* (left) and CSI in a single Bayesian network (right) to represent the fact that variables $X$ and $\{Z, W\}$ are strongly conditionally independent given the *context* $Y = 0$, but not when $Y = 1$, in the CPT in Table 1.

Boutilier et al. [1] formalized the notion of contextual strong independence with *context-strong independence* (CSI). (CSI is called *context-specific independence* in [1].) Let $\mathbf{X}, \mathbf{Y}, \mathbf{C}, \mathbf{Z}$ be pairwise disjoint sets of variables. We say $\mathbf{X}$ and $\mathbf{Z}$ are *strongly independent* given $\mathbf{Y}$ and the *context* $\mathbf{C} = \mathbf{c}$, denoted $I(\mathbf{X} \perp \mathbf{Z} \mid \mathbf{Y}, \mathbf{C} = \mathbf{c})$, if:

$$P(\mathbf{X} \mid \mathbf{Y}, \mathbf{C} = \mathbf{c}, \mathbf{Z}) = P(\mathbf{X} \mid \mathbf{Y}, \mathbf{C} = \mathbf{c}), \quad (3)$$

whenever $P(\mathbf{Y}, \mathbf{C} = \mathbf{c}, \mathbf{Z}) > 0$. It should be clear from equation (3) that strong CI is a *special case* of CSI, namely, CSI becomes strong CI when the CSI holds for all $\mathbf{c} \in V_{\mathbf{C}}$. That is, if the context-strong independence of $\mathbf{X}$ and $\mathbf{Z}$ given $\mathbf{Y}$ and $\mathbf{C} = \mathbf{c}$ holds for all $\mathbf{c} \in V_{\mathbf{C}}$, then we simply say $\mathbf{X}$ and $\mathbf{Z}$ are conditionally independent given $\mathbf{YC}$. Instead of using Bayesian multinets to capture contextual strong independencies, Boutilier et al. [1] represent these contextual strong statements with a single Bayesian network by introducing auxiliary nodes. Given the CPT in Table 1, the contextual strong independence of variables $X$ and $\{Z, W\}$ given $Y = 0$, and dependence when $Y = 1$, is captured with auxiliary nodes such as $X_{Y=0}$ as shown in Figure 1 (right).

In [2], Galles and Pearl studied the notion of *probabilistic causal irrelevance* (PCI). We say $\mathbf{Z}$ is *probabilistically causally irrelevant* to $\mathbf{X}$ given $\mathbf{Y}$ if for a *fixed* $\mathbf{y} \in V_{\mathbf{Y}}$, the following relationship holds:

$$P(\mathbf{X} \mid \mathbf{Y} = \mathbf{y}, \mathbf{Z}) = P(\mathbf{X} \mid \mathbf{Y} = \mathbf{y}). \quad (4)$$

Thus, PCI tries to capture the same contextual strong independencies as defined by equation (3). It follows that if the causal irrelevance holds for all $\mathbf{y} \in V_{\mathbf{Y}}$, then $\mathbf{X}$ and $\mathbf{Z}$ are probabilistically conditionally independent given $\mathbf{Y}$ in the usual sense.

It should be clear that the notions of ASI, CSI and PCI all try to capture contextual strong independence, where the partition of the CPT is defined by $\mathbf{Y} = \mathbf{y}$. In the next section, however, we introduce a *weak* contextual independence in which the partition of the CPT is *not* explicitly defined by the context $\mathbf{Y} = \mathbf{y}$.

## 3  CONTEXTUAL WEAK INDEPENDENCE (CWI)

We now introduce the notion of *contextual weak independence* (CWI), namely, variables $\mathbf{X}$ and $\mathbf{Z}$ are *weakly* independent given the *context* $\mathbf{Y} = \mathbf{y}$. We show that this contextual independence is more general than the notions of ASI, CSI, and PCI.

The notions of ASI, CSI and PCI are too strong to capture a weaker form of independence. Let $P$ be a joint probability distribution over the set of variables $\mathbf{U}$, where $X, Y, Z, W \in \mathbf{U}$. Let $V_X = \{0, 1, 2\}$, $V_Y = \{0, 1\}$ and $V_Z = V_W = \{0, 1, 2, 3\}$. Consider the CPT shown in Table 2, where configurations $\mathbf{c}$ with $P(\mathbf{c}) = 0$ are not shown. By equation (1), $X$ and $\{Z, W\}$ are *not* conditionally independent given $Y$. More importantly, there is *no* contextual strong independency between variables $X$ and $\{Z, W\}$ given the context $Y = 0$ or $Y = 1$. For example, if $Y = 0$, then

$$P(X = 0 \mid Y = 0, Z = 0, W = 0)$$
$$= p_1 \neq P(X = 0 \mid Y = 0, Z = 0, W = 3) = 0, \quad (5)$$

and if $Y = 1$, then

$$P(X = 0 \mid Y = 1, Z = 0, W = 0)$$
$$= p_5 \neq P(X = 0 \mid Y = 1, Z = 0, W = 1) = 0. \quad (6)$$

By equations (5) and (6), the notion of ASI does *not* capture any form of independence between variables $X$ and $\{Z, W\}$ given $Y = 0$ or $Y = 1$ as reflected by the Bayesian multinets depicted in Figure 2 (left). Furthermore, equations (5) and (6) also indicate that variables $X$ and $\{Z, W\}$ are not *contextually independent* given either the *context* $Y = 0$ or the *context* $Y = 1$. Consequently, the Bayesian network in Figure 2 (right) using the notion of CSI by incorporating auxiliary nodes does *not* reflect any independence between $X$ and $\{Z, W\}$ given $Y = 0$ or $Y = 1$. Figure 2 clearly illustrates that the notions of ASI and CSI do *not* capture any independency between variables $X$ and $\{Z, W\}$ given $Y = 0$. (Equations (5) and (6) also indicate that variables $\{Z, W\}$ are not *causally irrelevant* to $X$ when $Y = 0$ or $Y = 1$.)

Table 2: Variables $X$ and $\{Z, W\}$ are *weakly independent* given the *context* $Y = 0$, but not when $Y = 1$.

|       | X | Y | Z | W | $P(X \mid Y, Z, W)$ |
|-------|---|---|---|---|---|
| $t_1$ | 0 | 0 | 0 | 0 | $p_1$ |
| $t_2$ | 0 | 0 | 0 | 1 | $p_1$ |
| $t_3$ | 0 | 0 | 1 | 0 | $p_1$ |
| $t_4$ | 0 | 0 | 1 | 1 | $p_1$ |
| $t_5$ | 1 | 0 | 0 | 0 | $p_2$ |
| $t_6$ | 1 | 0 | 0 | 1 | $p_2$ |
| $t_7$ | 1 | 0 | 1 | 0 | $p_2$ |
| $t_8$ | 1 | 0 | 1 | 1 | $p_2$ |
| $t_9$ | 2 | 0 | 2 | 2 | $p_3$ |
| $t_{10}$ | 2 | 0 | 2 | 3 | $p_3$ |
| $t_{11}$ | 2 | 0 | 3 | 2 | $p_4$ |
| $t_{12}$ | 2 | 0 | 3 | 3 | $p_4$ |
| $t_{13}$ | 0 | 1 | 0 | 0 | $p_5$ |
| $t_{14}$ | 1 | 1 | 0 | 2 | $p_6$ |
| $t_{15}$ | 2 | 1 | 1 | 3 | $p_7$ |

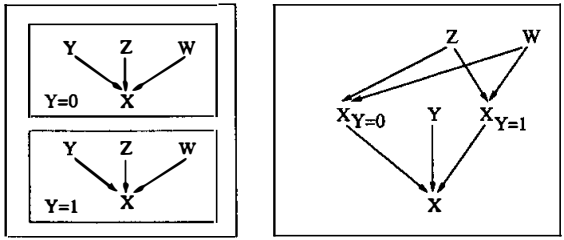

Figure 2: The notions of ASI in Bayesian multinets (left) and CSI (right) do *not* indicate any independence between variables $X$ and $\{Z, W\}$ given $Y = 0$ in the CPT in Table 2.

We now informally demonstrate a *weak* independence between variables $X$ and $\{Z, W\}$ given the context $Y = 0$. Let us focus on the configurations $(x, Y = 0, z, w)$ in Table 2, namely, the set of configurations $\{t_1, t_2, \ldots, t_{12}\}$. Recall that the domain of variables $X$, $Y$, $Z$, and $W$ are $V_X = \{0, 1, 2\}$, $V_Y = \{0\}$, and $V_Z = V_W = \{0, 1, 2, 3\}$. We make a few observations. When $Y = 0$, only specific values $x \in V_X$ appear with specific values $z \in V_Z$ and $w \in V_W$. That is,

$$X = 0, 1 \iff Z = 0, 1 \text{ and } W = 0, 1,$$

and

$$X = 2 \iff Z = 2, 3 \text{ and } W = 2, 3.$$

In other words, when $X = 0$ then $Z$ is never 2 or 3 and so forth. Based on this observation let us partition the CPT in Table 2 into the three separate CPTs $\{t_1, t_2, \ldots, t_8\}$, $\{t_9, t_{10}, t_{11}, t_{12}\}$, and $\{t_{13}, t_{14}, t_{14}\}$. Consider the CPT $\{t_1, t_2, \ldots, t_8\}$. Let us define *new* domains $V'_X$, $V'_Y$, $V'_Z$, $V'_W$, based on the values that appear. We obtain

$$V'_X = \{0, 1\}, \; V'_Y = \{0\}, \; V'_Z = \{0, 1\}, \; V'_W = \{0, 1\}. \quad (7)$$

With respect to the new domains in equation (7), by definition variables $X$ and $\{Z, W\}$ are conditionally independent given $Y$ in the CPT $\{t_1, t_2, \ldots, t_8\}$. Now consider the CPT $\{t_9, t_{10}, t_{11}, t_{12}\}$. Similarly, we define new domains $V''_X$, $V''_Y$, $V''_Z$, $V''_W$, for the variables based on the values that appear. Hence,

$$V''_X = \{2\}, \; V''_Y = \{0\}, \; V''_Z = \{2, 3\}, \; V''_W = \{2, 3\}. \quad (8)$$

With respect to the domains in equation (8), however, variables $X$ and $\{Z, W\}$ are *not* conditionally independent given $Y$ since

$$P(X = 2 \mid Y = 0, Z = 2, W = 2)$$
$$= p_3 \neq P(X = 2 \mid Y = 0, Z = 3, W = 3) = p_4.$$

We now formalize this idea.

We begin by recalling some familiar notions about relations. Given a distribution $P$ over the set of variables $\mathbf{U} = \mathbf{XYZ}$, let

$$\mathbf{C} = \{ \, t = (\mathbf{x}, \mathbf{y}, \mathbf{z}) \mid P(\mathbf{x} \mid \mathbf{y}, \mathbf{z}) > 0 \, \}. \quad (9)$$

Given any subset $\mathbf{V} \subseteq \mathbf{U}$, we can define an equivalence relation $\theta(\mathbf{V})$ on $\mathbf{C}$: for all $t_i, t_j \in \mathbf{C}$,

$$t_i \; \theta(\mathbf{V}) \; t_j, \; \text{if } t_i[\mathbf{V}] = t_j[\mathbf{V}], \quad (10)$$

where $t_k[\mathbf{V}]$ denotes the value of $\mathbf{V}$ in the tuple $t_k$.

Consider two equivalence relations $\theta(\mathbf{V})$ and $\theta(\mathbf{W})$ on $T$, where $\mathbf{V}, \mathbf{W} \subseteq \mathbf{U}$. The binary operator $\circ$, called the *composition*, is defined by: for $t_i, t_k \in T$,

$$t_i \; \theta(\mathbf{V}) \circ \theta(\mathbf{W}) \; t_k, \quad (11)$$

if for some $t_j \in T$ both

$$t_i \; \theta(\mathbf{V}) \; t_j \text{ and } t_j \; \theta(\mathbf{W}) \; t_k.$$

Given two equivalence relations $\theta(\mathbf{V})$ and $\theta(\mathbf{W})$ on $T$, it can be shown that $\theta(\mathbf{V}) \circ \theta(\mathbf{W})$ is an equivalence relation (a partition) if and only if $\theta(\mathbf{V}) \circ \theta(\mathbf{W}) = \theta(\mathbf{W}) \circ \theta(\mathbf{V})$. We now define contextual weak independence (CWI) as follows.

Let $\mathbf{X}, \mathbf{Y}, \mathbf{Z}$ be pairwise disjoint sets of variables in $\mathbf{U}$. Let $\theta(\mathbf{XY} = \mathbf{y})$ and $\theta(\mathbf{Y} = \mathbf{yZ})$ be partitions on $\mathbf{C}$ in equation (9). We say variables $\mathbf{X}$ and $\mathbf{Z}$ are *weakly independent* given the *context* $\mathbf{Y} = \mathbf{y}$, denoted $WI(\mathbf{X} \perp \mathbf{Z}|\mathbf{Y} = \mathbf{y})$, if the following two conditions hold:

(i) $\theta(\mathbf{XY} = \mathbf{y}) \circ \theta(\mathbf{Y} = \mathbf{yZ}) = \theta(\mathbf{Y} = \mathbf{yZ}) \circ \theta(\mathbf{XY} = \mathbf{y})$, and

(ii) there *exists* an equivalence class $\pi$ in $\theta(\mathbf{XY} = \mathbf{y}) \circ \theta(\mathbf{Y} = \mathbf{yZ})$ such that: for any given $\mathbf{x} \in V_\mathbf{X}^\pi$, then for all $\mathbf{z} \in V_\mathbf{Z}^\pi$:

$$P(\mathbf{x} \mid \mathbf{y}, \mathbf{z}) = P(\mathbf{x} \mid \mathbf{y}), \quad (12)$$

where $V_\mathbf{X}^\pi, V_\mathbf{Z}^\pi$ are defined as:

$$V_\mathbf{X}^\pi = \{ \mathbf{x} \mid t = (\mathbf{X} = \mathbf{x}, \mathbf{Y} = \mathbf{y}, \mathbf{Z} = \mathbf{z}) \in \pi \}, \quad (13)$$

and

$$V_\mathbf{Z}^\pi = \{ \mathbf{z} \mid t = (\mathbf{X}, \mathbf{Y} = \mathbf{y}, \mathbf{Z} = \mathbf{z}) \in \pi \}. \quad (14)$$

Condition (i) says that the composite relation $\theta(\mathbf{XY} = \mathbf{y}) \circ \theta(\mathbf{Y} = \mathbf{yZ})$ is an equivalence relation. This is the necessary condition for $WI(\mathbf{X} \perp \mathbf{Z} \mid \mathbf{Y} = \mathbf{y})$ to hold. Condition (ii) says that variables $\mathbf{X}$ and $\mathbf{Z}$ are conditionally independent given $\mathbf{Y} = \mathbf{y}$ with respect to the new domains $V_\mathbf{X}^\pi$ and $V_\mathbf{Z}^\pi$.

For example, let us verify that variables $X$ and $\{Z, W\}$ are *weakly independent* given the *context* $Y = 0$ in the CPT shown in Table 2. By equation (9),

$$\mathbf{C} = \{ t_1, t_2, \ldots, t_{12} \}.$$

By equation (10), we obtain the following equivalence relations on $\mathbf{C}$:

$\theta(XY = 0)$
$= \{ \{t_1, t_2, t_3, t_4\}, \{t_5, t_6, t_7, t_8\}, \{t_9, t_{10}, t_{11}, t_{12}\} \}$,

and

$\theta(Y = 0ZW) = \{ \{t_1, t_5\}, \{t_2, t_6\}, \{t_3, t_7\}, \{t_4, t_8\},$
$\{t_9\}, \{t_{10}\}, \{t_{11}\}, \{t_{12}\} \}.$

Applying equation (11), we obtain:

$\theta(XY = 0) \circ \theta(Y = 0ZW)$
$= \{ \pi_1 = \{t_1, t_2, \ldots, t_8\}, \pi_2 = \{t_9, t_{10}, t_{11}, t_{12}\} \}$
$= \theta(Y = 0ZW) \circ \theta(XY = 0). \quad (15)$

Applying equations (13) and (14) on the configurations $\pi_1 = \{t_1, t_2, \ldots, t_8\}$, we obtain

$V_X^{\pi_1} = \{ x \mid t = (X = x, Y = y, Z, W) \in \pi_1 \}$
$= \{0, 1\}, \quad (16)$

and

$V_{ZW}^{\pi_1}$
$= \{ (z, w) \mid t = (X, Y = y, Z = z, W = w) \in \pi_1 \}$
$= \{(0,0), (0,1), (1,0), (1,1)\}. \quad (17)$

It can be verified that variables $X$ and $\{Z, W\}$ are conditionally independent given $Y = 0$ in equivalence class $\pi_1$ with respect to the new domains $V_X^{\pi_1} = \{0, 1\}$ and $V_{ZW}^{\pi_1} = \{(0,0), (0,1), (1,0), (1,1)\}$. This process can be repeated for the set of configurations $\pi_2 = \{t_9, t_{10}, t_{11}, t_{12}\}$. The computed new domains are $V_X^{\pi_2} = \{2\}$ and $V_{ZW}^{\pi_2} = \{(2,2), (2,3), (3,2), (3,3)\}$. As already mentioned, however,

$P(X = 2 \mid Y = 0, Z = 2, W = 2)$
$= p_3 \neq P(X = 2 \mid Y = 0, Z = 3, W = 3) = p_4.$

Thus, variables $X$ and $\{Z, W\}$ are *not* conditionally independent given $Y = 0$ with respect to the new domains in equivalence class $\pi_2$. However, condition (ii) is still satisfied since the conditional independence holds for at least one equivalence class, i.e., $\pi_1 = \{t_1, t_2, \ldots, t_8\}$. By definition, variables $X$ and $\{Z, W\}$ are weakly independent given the context $Y = 0$.

The conditional independence in equivalence class $\pi_1$ is reflected by the Bayesian network in Figure 3 (left). The dependency in equivalence class $\pi_2$ is reflected by the Bayesian network in Figure 3 (middle). As with ASI and CSI in Figure 2 (left, right), variables $X$ and $\{Z, W\}$ are *not* weakly independent given the context $Y = 1$ as shown in Figure 3 (right).

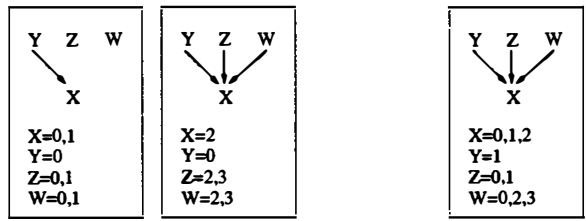

Figure 3: Unlike the various Bayesian networks in Figure 2, the Bayesian network (left) can capture the *weak independence* of variables $X$ and $\{Z, W\}$ given the *context* $Y = 0$ in the CPT in Table 3.

The important point is that the notion of contextual weak independence (CWI) can capture a more general form of independence than the notions of contextual strong independence. In the above example, the contextual strong notions of ASI and CSI do *not* capture any independence between variables $X$ and $\{Z, W\}$ given $Y = 0$ as shown by the various Bayesian networks in Figure 2. On the other hand, the notion of variables $X$ and $Z$ being *weakly independent* given the context $Y = 0$ is captured by the Bayesian network in Figure 3 (left).

## 4 WEAK INDEPENDENCE (WI)

In this section, we show another important distinction between the notion of contextual weak independence and the contextual strong notions of ASI, CSI and PCI. If the contextual strong independence of $X$ and $Z$ holds for *all* contexts $Y = y$, $y \in V_Y$, then the notions of ASI, CSI and PCI become strong CI, namely, $X$ and $Z$ are strongly conditional independent given $Y$. This is *not* necessarily the case with contextual weak independence. If $X$ and $Z$ are weakly independent for all contexts $Y = y$, $y \in V_Y$, then we say $X$ and $Z$ are *weakly independent* given $Y$. Just as contextual weak independence (CWI) is more general than the notions of contextual strong independence (CSI, ASI, PCI), we show that the notion of *weak independence* (WI) is a more general noncontextual independency than strong CI. In the next section, we show that the class of WI and CI together has a *complete* axiomatization.

Let $\mathbf{X}, \mathbf{Y}, \mathbf{Z}$ be pairwise disjoint sets of variables in $\mathbf{U}$. Let $\theta(\mathbf{XY})$ and $\theta(\mathbf{YZ})$ be partitions on $\mathbf{C}$ in equation (9). We say variables $\mathbf{X}$ and $\mathbf{Z}$ are *weakly independent* given $\mathbf{Y}$, denoted $WI(\mathbf{X} \perp \mathbf{Z} \mid \mathbf{Y})$, if the following two conditions hold:

$$(i) \quad \theta(\mathbf{XY}) \circ \theta(\mathbf{YZ}) = \theta(\mathbf{YZ}) \circ \theta(\mathbf{XY}), \quad (18)$$

and (ii) for *every* equivalence class $\pi$ in the equivalence relation $\theta(\mathbf{YZ}) \circ \theta(\mathbf{XY})$, if given any $\mathbf{x} \in V_\mathbf{X}^\pi$, $\mathbf{y} \in V_\mathbf{Y}^\pi$, then for all $\mathbf{z} \in V_\mathbf{Z}^\pi$:

$$P(\mathbf{x} \mid \mathbf{y}, \mathbf{z}) = P(\mathbf{x} \mid \mathbf{y}), \quad (19)$$

where the *new* domains $V_\mathbf{X}^\pi$, $V_\mathbf{Y}^\pi$ and $V_\mathbf{Z}^\pi$ are defined as:

$$V_\mathbf{X}^\pi = \{ \mathbf{x} \mid t = (\mathbf{X} = \mathbf{x}, \mathbf{Y}, \mathbf{Z}) \in \pi \}, \quad (20)$$
$$V_\mathbf{Y}^\pi = \{ \mathbf{y} \mid t = (\mathbf{X}, \mathbf{Y} = \mathbf{y}, \mathbf{Z}) \in \pi \}, \quad (21)$$
$$V_\mathbf{Z}^\pi = \{ \mathbf{y} \mid t = (\mathbf{X}, \mathbf{Y}, \mathbf{Z} = \mathbf{z}) \in \pi \}. \quad (22)$$

For example, consider the CPT shown in Table 3, where $V_X = V_Z = V_W = \{0, 1, 2, 3\}$ and $V_Y = \{0, 1\}$. Variables $X$ and $\{Z, W\}$ are *not* conditionally independent given $Y$. However, variables $X$ and $\{Z, W\}$ are *weakly independent* given $Y$. By equation (9), $\mathbf{C} = \{t_1, t_2, \ldots, t_{32}\}$. By equation (10), we obtain the equivalence relations $\theta(XY)$ and $\theta(YZW)$ on $\mathbf{C}$:

$\theta(XY) =$

$\{ \{t_1, t_2, t_3, t_4\}, \{t_5, t_6, t_7, t_8\}, \{t_9, t_{10}, t_{11}, t_{12}\},$
$\{t_{13}, t_{14}, t_{15}, t_{16}\}, \{t_{17}, t_{18}, t_{19}, t_{20}\}, \{t_{21}, t_{22}, t_{23}, t_{24}\},$
$\{t_{25}, t_{26}, t_{27}, t_{28}\}, \{t_{29}, t_{30}, t_{31}, t_{32}\} \},$

Table 3: Variables $X$ and $\{Z, W\}$ are *weakly independent* given $Y$.

|  | X | Y | Z | W | $P(X\mid Y,Z,W)$ |
|---|---|---|---|---|---|
| $t_1$ | 0 | 0 | 0 | 0 | $p_1$ |
| $t_2$ | 0 | 0 | 0 | 1 | $p_1$ |
| $t_3$ | 0 | 0 | 1 | 0 | $p_1$ |
| $t_4$ | 0 | 0 | 1 | 1 | $p_1$ |
| $t_5$ | 1 | 0 | 0 | 0 | $p_1$ |
| $t_6$ | 1 | 0 | 0 | 1 | $p_1$ |
| $t_7$ | 1 | 0 | 1 | 0 | $p_1$ |
| $t_8$ | 1 | 0 | 1 | 1 | $p_1$ |
| $t_9$ | 2 | 0 | 2 | 2 | $p_1$ |
| $t_{10}$ | 2 | 0 | 2 | 3 | $p_1$ |
| $t_{11}$ | 2 | 0 | 3 | 2 | $p_1$ |
| $t_{12}$ | 2 | 0 | 3 | 3 | $p_1$ |
| $t_{13}$ | 3 | 0 | 2 | 2 | $p_1$ |
| $t_{14}$ | 3 | 0 | 2 | 3 | $p_1$ |
| $t_{15}$ | 3 | 0 | 3 | 2 | $p_1$ |
| $t_{16}$ | 3 | 0 | 3 | 3 | $p_1$ |
| $t_{17}$ | 0 | 1 | 0 | 0 | $p_2$ |
| $t_{18}$ | 0 | 1 | 0 | 3 | $p_2$ |
| $t_{19}$ | 0 | 1 | 3 | 0 | $p_2$ |
| $t_{20}$ | 0 | 1 | 3 | 3 | $p_2$ |
| $t_{21}$ | 2 | 1 | 0 | 0 | $p_2$ |
| $t_{22}$ | 2 | 1 | 0 | 3 | $p_2$ |
| $t_{23}$ | 2 | 1 | 3 | 0 | $p_2$ |
| $t_{24}$ | 2 | 1 | 3 | 3 | $p_2$ |
| $t_{25}$ | 1 | 1 | 1 | 1 | $p_3$ |
| $t_{26}$ | 1 | 1 | 1 | 2 | $p_3$ |
| $t_{27}$ | 1 | 1 | 2 | 1 | $p_3$ |
| $t_{28}$ | 1 | 1 | 2 | 2 | $p_3$ |
| $t_{29}$ | 3 | 1 | 1 | 1 | $p_4$ |
| $t_{30}$ | 3 | 1 | 1 | 2 | $p_4$ |
| $t_{31}$ | 3 | 1 | 2 | 1 | $p_4$ |
| $t_{32}$ | 3 | 1 | 2 | 2 | $p_4$ |

and

$\theta(YZW) =$

$\{ \{t_1, t_5\}, \{t_2, t_6\}, \{t_3, t_7\}, \{t_4, t_8\}, \{t_9, t_{13}\}, \{t_{10}, t_{14}\},$
$\{t_{11}, t_{15}\}, \{t_{12}, t_{16}\}, \{t_{17}, t_{21}\}, \{t_{18}, t_{22}\}, \{t_{19}, t_{23}\},$
$\{t_{20}, t_{24}\}, \{t_{25}, t_{29}\}, \{t_{26}, t_{30}\}, \{t_{27}, t_{31}\}, \{t_{28}, t_{32}\} \}.$

Applying equation (11), we obtain:

$\theta(XY) \circ \theta(YZW)$
$= \{ \pi_1 = \{t_1, t_2, t_3, t_4, t_5, t_6, t_7, t_8\},$
$\quad \pi_2 = \{t_9, t_{10}, t_{11}, t_{12}, t_{13}, t_{14}, t_{15}, t_{16}\},$
$\quad \pi_3 = \{t_{17}, t_{18}, t_{19}, t_{20}, t_{21}, t_{22}, t_{23}, t_{24}\},$
$\quad \pi_4 = \{t_{25}, t_{26}, t_{27}, t_{28}, t_{29}, t_{30}, t_{31}, t_{32}\} \}$
$= \theta(YZW) \circ \theta(XY). \quad (23)$

Condition (i) of the definition of WI is then satisfied. With respect to the equivalence class $\pi_1 =$

$\{t_1, t_2, \ldots, t_8\}$, we obtain by equations (20)-(22) the *new* domains $V_X^{\pi_1}$, $V_Y^{\pi_1}$ and $V_{ZW}^{\pi_1}$:

$$V_X^{\pi_1} = \{0,1\}, \quad V_Y^{\pi_1} = \{0\},$$
$$V_{ZW}^{\pi_1} = \{(0,0), (0,1), (1,0), (1,1)\}.$$

With respect to the new domains $V_X^{\pi_1}$, $V_Y^{\pi_1}$ and $V_{ZW}^{\pi_1}$, variables $X$ and $\{Z,W\}$ are conditionally independent given $Y$. This conditional independence can be verified similarly in the other equivalence classes $\pi_2 = \{t_9, t_{10}, \ldots, t_{16}\}$, $\pi_3 = \{t_{17}, t_{18}, \ldots, t_{24}\}$, and $\pi_4 = \{t_{25}, t_{26}, \ldots, t_{32}\}$. By definition, variables $X$ and $\{Z,W\}$ are weakly independent given $Y$.

There are two equivalent ways to view the representation of WI. One is in terms of multiple Bayesian networks as in [3]. The difference here is that each Bayesian network has the *same* dependency structure as shown in Figure 4. As with the Bayesian multinets approach [3], the particular CPT to be used is defined by the evidence. An alternative viewpoint is to represent WI in a single Bayesian network as with the notion of CSI [1]. The difference here is that no artificial nodes need to be incorporated into the Bayesian network. Instead each node is simply associated with a *set* of CPTs compared to a single CPT as with the notion of CI in traditional Bayesian networks.

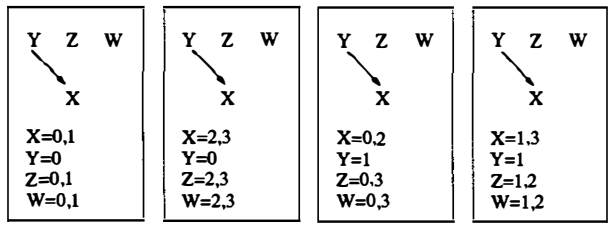

Figure 4: The use of multiple Bayesian networks with the same structure to represent the *weak independence* of variables $X$ and $\{Z,W\}$ given $Y$ in the CPT in Table 3.

In the next section, we turn our attention to developing an axiomatic basis for WI.

## 5 COMPLETE AXIOMATIZATION FOR WI AND CI

There are both theoretical and practical reasons for developing a complete axiomatization for a new type of dependency. When designing a probabilistic network [6], it is always desirable that the designer have complete knowledge about the given input dependencies and all of their logical consequences. On the practical side, in the absence of knowing a logical consequence, an inference engine may spend precious time on derivations bearing no relevance to the task at hand.

In [2], Galles and Pearl develop axioms to formalize the notion of probabilistic causal irrelevance (PCI) in a similar spirit to the semi-graphoid axioms [6] for CI. As shown in Section 3, however, the contextual strong notion of PCI is too restrictive to capture contextual weak independence. In this section, we study *complete* axiomatizations for both the class of *weak independence* (WI) and the class of CI and WI together. Note that we only consider CI and WI statements defined with respect to the same fixed set of variables called *nonembedded* (full) independencies. That is, we do not consider axioms that involve statements involving a mixture of variables such as Pearl's *contraction* axiom (3.6d) [6].

**Theorem 1** Let $\mathbf{X}, \mathbf{Y}, \mathbf{Z}, \mathbf{W}$ be subsets of variables from $\mathbf{U}$ such that $\mathbf{XYZ} = \mathbf{U}$. A *complete* set of inference axioms for WI is:

WI1: $WI(\mathbf{X} \perp \mathbf{U} - \mathbf{Y} \mid \mathbf{Y})$, $\mathbf{X} \subseteq \mathbf{Y}$;

WI2: $WI(\mathbf{X} \perp \mathbf{U} - \mathbf{XY} \mid \mathbf{Y}) \Rightarrow$
$WI(\mathbf{X} - \mathbf{W} \perp \mathbf{U} - \mathbf{X}(\mathbf{Y} - \mathbf{W}) \mid \mathbf{Y})$,
$WI(\mathbf{XW} \perp \mathbf{U} - \mathbf{XYW} \mid \mathbf{Y})$, $\mathbf{W} \subseteq \mathbf{Y}$;

WI3: $WI(\mathbf{X} \perp \mathbf{U} - \mathbf{XY} \mid \mathbf{Y}) \Rightarrow$
$WI(\mathbf{X} \perp \mathbf{U} - \mathbf{XYW}) \mid \mathbf{YW})$.

WI1, WI2, and WI3 are called *reflexivity*, *transport*, and *augmentation* axioms, respectively. It is interesting to note that there is no transitivity axiom for WI.

**Theorem 2** Let $\mathbf{X}, \mathbf{Y}, \mathbf{Z_1}, \mathbf{Z_2}$ be pairwise disjoint subsets of variables from $\mathbf{U}$ such that $\mathbf{XYZ_1Z_2} = \mathbf{U}$. A *complete* set of inference axioms for CI and WI together is:

CI &WI1: $I(\mathbf{X} \perp \mathbf{U} - \mathbf{XY} \mid \mathbf{Y}) \Rightarrow$
$WI(\mathbf{Y} \perp \mathbf{U} - \mathbf{XY}) \mid \mathbf{Y})$,

CI &WI2: $WI(\mathbf{X} \perp \mathbf{Z_2} \mid \mathbf{YZ_1})$, $WI(\mathbf{X} \perp \mathbf{Z_1} \mid \mathbf{YZ_2})$,
$I(\mathbf{Z_1} \perp \mathbf{Z_2} \mid \mathbf{YX}) \Rightarrow WI(\mathbf{X} \perp \mathbf{Z_1Z_2} \mid \mathbf{Y})$.

CI &WI1 and CI &WI2 are called *weaken* and *transitivity* axioms, respectively.

## 6 GRANULAR PROBABILISTIC NETWORKS

In this section, we present a brief discussion on granular probabilistic networks [5]. We use the term granular to mean the ability to *coarsen* and *refine* parts of a probabilistic network. We will show that WI is a necessary and sufficient condition for ensuring *consistency* in such granular networks.

Koller and Pfeffer [5] suggested a framework for the modeling of and inference in a large Bayesian network. In this framework, parts of the network can be coarsened and refined. For example, consider the Bayesian network for a car accident [5] as shown in Figure 5. One can *refine* the node $Car$ to reveal the internal structure as shown in Figure 6, where the rest of the network is not illustrated due to space limitations.

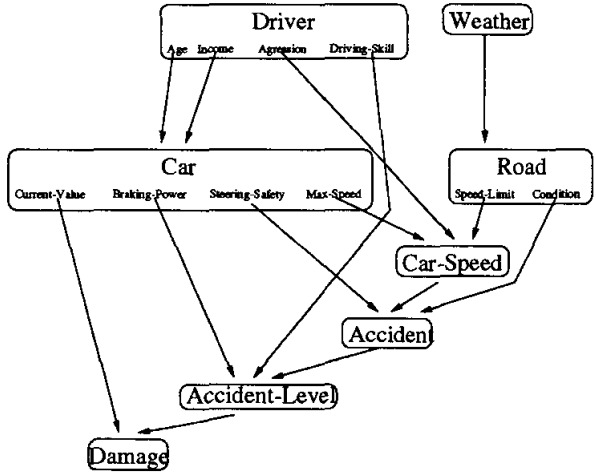

Figure 5: A Bayesian network for a car accident.

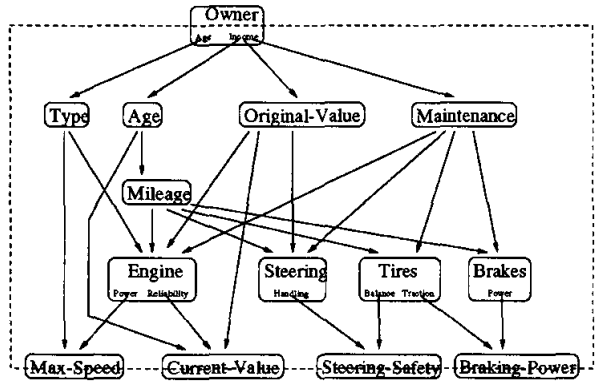

Figure 6: Refining the Bayesian network in Figure 5 to reveal the internal structure of the variable $Car$.

Our purpose here is to present two operators NEST and UNNEST for coarsening and refining a probabilistic network. The main result of this section is that WI is a necessary and sufficient condition for ensuring *consistency* in such granular networks. (The notion of CI is only a sufficient condition.)

We introduce the operator NEST to *coarsen* a joint probability distribution. The NEST operator, denoted $N$, is used to coarsen parts of a network. Intuitively, $N_{B=Y}(P_{XY})$ groups together all the $Y$-values into a nested distribution given the same $X$-value.

In the following definitions, we use boldface letters such as $t$ to denote an entire configuration $c$ along with $P(c)$, i.e., $t = (c, P(c))$. Recall $t[X]$ denotes the value of $X$ in configuration $c$. The straightforward idea of Nest and Unnest involve rather cumbersome definitions. The examples that follow should clarify any confusion.

Coarsening variables $Y$ in distribution $P$ as attribute $B$ is the distribution $N_{B=Y}(P)$ with variables $X, B, P(X, B)$ defined by

$$N_{B=Y}(P) = \{t \mid t[X] = u[X] \text{ and } t[B] = \{u[Y, P]\},$$
$$\text{and } t[P(X, B)] = \sum_u u[P]\},$$

where $u \in P$. The attribute $P$ in the value of $B$ is relabelled $P(Y)$ and the values normalized.

For example, consider the distribution in Figure 7. Coarsening variables $\{A_2, A_3\}$ as $B$ is the distribution $N_{B=\{A_2,A_3\}}(P)$ depicted in Figure 8.

| $A_1$ | $A_2$ | $A_3$ | $P(A_1, A_2, A_3)$ |
|---|---|---|---|
| 1 | 1 | 2 | 0.125 |
| 1 | 3 | 4 | 0.250 |
| 1 | 5 | 6 | 0.125 |
| 2 | 1 | 3 | 0.125 |
| 2 | 2 | 4 | 0.125 |
| 3 | 0 | 0 | 0.125 |
| 3 | 0 | 1 | 0.125 |

Figure 7: A joint probability distribution $P$ over $U = \{A_1, A_2, A_3\}$.

We now introduce the operator UNNEST to *refine* parts of the network. The UNNEST operator, denoted $U$, reveals the nested variables. Intuitively, $U_{B=Y}(P)$ joins each $X$-value with each tuple in the corresponding B-value.

Revealing the nested variables $Y$ in attribute $B$ of $P_{XB}$ is the distribution $U_{B=Y}(P)$ with variables $XY, P(X, Y)$ defined by

$$U_{B=Y}(P) = \{t \mid t[X] = u[X] \text{ and } t[Y] \in u[B]$$
$$\text{and } t[P] = \sum_v v[P(XB)] \cdot w[P(Y)]\},$$

where $u \in P, v[X] = u[X], w \in v[B]$ and $w[Y] = t[Y]$. Note that we may write $U_{B=Y}(P)$ as $U_B(P)$ since $Y$ is implicitly implied by $B$.

For example, revealing the nested variables in coarsened distribution $P'$ depicted in Figure 8 results in the refined distribution $U_B(P')$ shown in Figure 7.

| $A_1$ | $B$ | | | $P(A_1, B)$ |
|---|---|---|---|---|
| 1 | $A_2$ | $A_3$ | $P(A_2, A_3)$ | 0.50 |
| | 1 | 2 | 0.25 | |
| | 3 | 4 | 0.50 | |
| | 5 | 6 | 0.25 | |
| 2 | $A_2$ | $A_3$ | $P(A_2, A_3)$ | 0.25 |
| | 1 | 3 | 0.5 | |
| | 2 | 4 | 0.5 | |
| 3 | $A_2$ | $A_3$ | $P(A_2, A_3)$ | 0.25 |
| | 0 | 0 | 0.5 | |
| | 0 | 1 | 0.5 | |

Figure 8: The coarser distribution $N_{B=\{A_2, A_3\}}(P)$ obtained by nesting variables $\{A_2, A_3\}$ in the distribution $P$ in Figure 7 as variable $B$.

The following example demonstrates that inconsistency may arise due to the fact that the NEST operator is not commutative. (In contrast, the UNNEST operator is commutative.) Consider the distribution $P$ in Figure 9.

| $A_1$ | $A_2$ | $A_3$ | $P(A_1, A_2, A_3)$ |
|---|---|---|---|
| 0 | 0 | 0 | 0.4 |
| 1 | 0 | 0 | 0.4 |
| 1 | 0 | 1 | 0.2 |

Figure 9: The joint probability distribution $P$ on $\mathbf{U} = \{A_1, A_2, A_3\}$.

Suppose we wish to coarsen the variables $A_1$ and $A_3$. Nesting $A_3$ as $B_3$ followed by $A_1$ as $B_1$ results in the distribution $N_{B_1=\{A_1\}}(N_{B_3=\{A_3\}}(P))$ depicted in Figure 10. On the other hand, coarsening $A_1$ as $B_1$ followed by $A_3$ as $B_3$ results in the distribution $N_{B_3=\{A_3\}}(N_{B_1=\{A_1\}}(P))$ illustrated in Figure 11.

| $B_1$ | $A_2$ | $B_3$ | | $P(B_1, A_2, B_3)$ |
|---|---|---|---|---|
| $A_1$ $P(A_1)$ | 0 | $A_3$ $P(A_3)$ | | 0.6 |
| 0　1.0 | | 0　1.0 | | |
| $A_1$ $P(A_1)$ | 0 | $A_3$ $P(A_3)$ | | 0.4 |
| 1　1.0 | | 0　0.66 | | |
| | | 1　0.33 | | |

Figure 10: The coarsened distribution $N_{B_1=\{A_1\}}(N_{B_3=\{A_3\}}(P))$, where $P$ is the joint probability distribution $P$ in Figure 9.

| $B_1$ | $A_2$ | $B_3$ | | $P(B_1, A_2, B_3)$ |
|---|---|---|---|---|
| $A_1$ $P(A_1)$ | 0 | $A_3$ $P(A_3)$ | | 0.8 |
| 0　0.5 | | 0　1.0 | | |
| 1　0.5 | | | | |
| $A_1$ $P(A_1)$ | 0 | $A_3$ $P(A_3)$ | | 0.2 |
| 1　1.0 | | 0　1.0 | | |

Figure 11: The coarsened distribution $N_{B_3=\{A_3\}}(N_{B_1=\{A_1\}}(P))$, where $P$ is the joint probability distribution $P$ in Figure 9.

We now show that WI is a necessary and sufficient condition for NEST to commute. In other words, WI ensures *consistency* in granular networks.

**Theorem 3** Let $P$ be a joint probability distribution over $\mathbf{U}$, and $\mathbf{X}$, $\mathbf{Y}$, $\mathbf{Z}$ pairwise disjoint subsets such that $\mathbf{XYZ} = \mathbf{U}$. $\mathbf{X}$ and $\mathbf{Z}$ are weakly independent given $\mathbf{Y}$ if and only if

$$N_{B_1=\mathbf{X}}(N_{B_2=\mathbf{Z}}(P)) = N_{B_2=\mathbf{Z}}(N_{B_1=\mathbf{X}}(P)).$$

*Proof:* ($\Rightarrow$) Suppose $\mathbf{X}$ and $\mathbf{X}$ are weakly independent given $\mathbf{Y}$. By definition, $\mathbf{X}$ and $\mathbf{Z}$ are conditionally independent given $\mathbf{Y}$ in each equivalence class in the equivalence relation $\theta(\mathbf{XY}) \circ \theta(\mathbf{YZ})$. One equivalence class is shown in Figure 12, where $c = a_1 + a_2 = b_1 + b_2$ by equation (2).

| $\mathbf{X}$ | $\mathbf{Y}$ | $\mathbf{Z}$ | $P(\mathbf{XYZ})$ |
|---|---|---|---|
| $x_1$ | $y_1$ | $z_1$ | $(a_1 b_1)/c$ |
| $x_1$ | $y_1$ | $z_2$ | $(a_1 b_2)/c$ |
| $x_2$ | $y_1$ | $z_1$ | $(a_2 b_1)/c$ |
| $x_2$ | $y_1$ | $z_2$ | $(a_2 b_2)/c$ |

Figure 12: $\mathbf{X}$ and $\mathbf{Z}$ are conditionally independent given $\mathbf{Y}$ in each equivalence class in the equivalence relation $\theta(\mathbf{XY}) \circ \theta(\mathbf{YZ})$.

It is clear that when $\mathbf{X}$ and $\mathbf{Z}$ are weakly independent given $\mathbf{Y}$, computing $N_{B_2=\mathbf{Z}}(P)$ only groups together tuples in the same equivalence class in the equivalence relation $\theta(\mathbf{XY}) \circ \theta(\mathbf{YZ})$. Computing $N_{B_2=\mathbf{Z}}(P)$ modifies the equivalence class in Figure 12 into the one depicted in Figure 13. The expressions for probability values of $B_2$ can be simplified since, for example

$$\frac{(a_1 b_1)c}{(a_1 b_1)c + (a_1 b_2)c} = \frac{b_1}{b_1 + b_2}.$$

Since the $B_2$-values are identical, computing $N_{B_1=\mathbf{X}}(N_{B_2=\mathbf{Z}}(P))$ modifies the equivalence class in Figure 13 into the one illustrated in Figure 14.

| X | Y | $B_2$ | | $P(X,Y,B_2)$ |
|---|---|---|---|---|
| $x_1$ | $y_1$ | Z | $P(Z)$ | $a_1(b_1+b_2)/c = a_1$ |
| | | $z_1$ | $(a_1b_1)/c = b_1$ | |
| | | $z_2$ | $(a_1b_2)/c = b_2$ | |
| $x_2$ | $y_1$ | Z | $P(Z)$ | $a_2(b_1+b_2)/c = a_2$ |
| | | $z_1$ | $(a_2b_1)/c = b_1$ | |
| | | $z_2$ | $(a_2b_2)/c = b_2$ | |

Figure 13: The distribution $N_{B_2=\mathbf{z}}(P)$.

| $B_1$ | | Y | $B_2$ | | $P(B_1,Y,B_2)$ |
|---|---|---|---|---|---|
| X | $P(X)$ | | Z | $P(Z)$ | |
| $x_1$ | $a_1$ | $y_1$ | $z_1$ | $b_1$ | $c$ |
| $x_2$ | $a_2$ | | $z_2$ | $b_2$ | |

Figure 14: The distribution $N_{B_1=\mathbf{x}}(N_{B_2=\mathbf{z}}(P))$.

This demonstrates that each equivalence class in $\theta(\mathbf{XY}) \circ \theta(\mathbf{YZ})$ is reduced into a single tuple with all the Z-values and X-values represented in one $B_1$-value and $B_2$-value, respectively. Since the conditional independence of X and Z given Y is symmetric, this argument can be applied to show that $N_{B_2=\mathbf{z}}(N_{B_1=\mathbf{x}}(P))$ also reduces the initial equivalence class in Figure 12 into the single tuple shown in Figure 14. This argument holds for each equivalence class in the equivalence relation $\theta(\mathbf{XY}) \circ \theta(\mathbf{YZ})$.

($\Leftarrow$) Let $P$ be a distribution such that $N_{B_1=\mathbf{x}}(N_{B_2=\mathbf{z}}(P)) = N_{B_2=\mathbf{z}}(N_{B_1=\mathbf{x}}(P))$ holds, where one tuple in $N_{B_1=\mathbf{x}}(N_{B_2=\mathbf{z}}(P))$ is depicted in Figure 14. The equality holding indicates that $c = a_1 + a_2 = b_1 + b_2$, as otherwise the order of nesting becomes relevant. The result of $U_{B_1}(N_{B_1=\mathbf{x}}(N_{B_2=\mathbf{z}}(P)))$ is depicted in Figure 13. It can be easily verified that X and Z are conditionally independent given Y in the normalized distribution $U_{B_2}(U_{B_1}(N_{B_1=\mathbf{x}}(N_{B_2=\mathbf{z}}(P))))$, shown in Figure 15. This argument holds for each tuple in $N_{B_1=\mathbf{x}}(N_{B_2=\mathbf{z}}(P))$. Therefore, X and Z are weakly independent given Y in the distribution $N_{B_1=\mathbf{x}}(N_{B_2=\mathbf{z}}(P))$. □

## 7  CONCLUSION

Many researchers including [1, 2, 3, 4, 6] have pointed out that the notion of (strong) CI is too restrictive to capture independencies that hold in some but not necessarily all contexts. This kind of contextual independency, referred to as *context-specific independence* (CSI) [1], *asymmetric independence* (ASI) [3], or *prob-

| X | Y | Z | $P(X,Y,Z)$ |
|---|---|---|---|
| $x_1$ | $y_1$ | $z_1$ | $b_1a_1/c$ |
| $x_1$ | $y_1$ | $z_2$ | $b_2a_1/c$ |
| $x_2$ | $y_1$ | $z_1$ | $b_1a_2/c$ |
| $x_2$ | $y_1$ | $z_2$ | $b_2a_2/c$ |

Figure 15: The distribution representing one tuple in $U_{B_2}(U_{B_1}(N_{B_1=\mathbf{x}}(N_{B_2=\mathbf{z}}(P))))$.

*abilistic causal irrelevance* (PCI) [2], can be used to facilitate the acquisition, representation, and inference of probabilistic knowledge.

In this paper, we suggest the use of a more general form of contextual independence called *contextual weak independence* (CWI) in Bayesian networks. It was explicitly demonstrated that CWI can detect independence that remains undetected by CSI, ASI and PCI. Furthermore, if the contextual strong independency holds for all contexts, then the notions of CSI, ASI and PCI become (strong) CI. On the other hand, if the contextual weak independency holds for all contexts, then CWI becomes *weak independence* (WI). Just as contextual weak independence (CWI) is more general than the notions of contextual strong independence, it was explicitly demonstrated that the notion of *weak independence* (WI) is a more general noncontextual independency than strong CI. We also studied *complete axiomatizations* for both the class of WI and the class of CI and WI together. Finally, the interesting property of WI being a necessary and sufficient condition for ensuring *consistency* in *granular probabilistic networks* [5] was demonstrated.